\documentclass[]{spie}  

\usepackage{amsmath,amsfonts,amssymb}
\usepackage{graphicx}
\usepackage{multirow}
\usepackage[colorlinks=true, allcolors=blue]{hyperref}

\title{From Single-Visit to Multi-Visit Image-Based Models: Single-Visit Models are Enough to Predict Obstructive Hydronephrosis}

\author[a,b]{Stanley Bryan Z. Hua}
\author[a,b]{Mandy Rickard}
\author[d]{John Weaver}
\author[d]{Alice Xiang}
\author[c]{Daniel Alvarez}
\author[c]{Kyla N Velear}
\author[c]{Kunj Sheth}
\author[d]{Gregory E. Tasian}
\author[a,b]{Armando J.Lorenzo}
\author[a,b]{Anna Goldenberg}
\author[a,b]{Lauren Erdman}
\affil[a]{University of Toronto, Toronto, Canada}
\affil[b]{Hospital for Sick Children, Toronto, Canada}
\affil[c]{Stanford Children's Health, Palo Alto, California}
\affil[d]{Children's Hospital of Philadelphia, Philadelphia, PA}

\authorinfo{Further author information: (Send correspondence to Stanley Bryan Hua or Lauren Erdman)\\ Stanley Bryan Hua E-mail: stanley.hua@mail.utoronto.ca \\ Lauren Erdman E-mail: lauren.erdman@sickkids.ca\\ } 

\begin{document} 
\maketitle

\begin{abstract}
Previous work has shown the potential of deep learning to predict renal obstruction using kidney ultrasound images. However, these image-based classifiers have been trained with the goal of single-visit inference in mind. We compare methods from video action recognition (i.e. convolutional pooling, LSTM, TSM) to adapt single-visit convolutional models to handle multiple visit inference. We demonstrate that incorporating images from a patient's past hospital visits provides only a small benefit for the prediction of obstructive hydronephrosis. Therefore, inclusion of prior ultrasounds is beneficial but prediction based on the latest ultrasound is sufficient for patient risk stratification. 
\end{abstract}


\section{INTRODUCTION}
\label{sec:intro}  

Hydronephrosis (HN) is a medical condition in which the urinary tract is dilated due to accumulation of urine. Postnatally, most cases of HN resolve with no intervention and no impact to renal function. Other times, HN may be caused by an underlying pathology that requires surgical treatment (termed obstructive hydronephrosis). Since spontaneous resolution may take longer than 3 years, infants diagnosed with HN require frequent clinical visits for ultrasound screens and invasive tests for renal function to monitor for obstruction. However, functional tests require painful procedures and risk exposing the infant to radiation. With the goal of reducing invasive testing in infants, researchers have demonstrated the potential of deep learning to predict renal obstruction from kidney ultrasounds alone \cite{hydro}. 

The current state of the art follows a 2D Siamese convolutional neural network (CNN) architecture \cite{hydro}. As input, it accepts two fixed-plane ultrasound images taken during a single unique hospital visit. The model extracts features from each image separately, and uses the concatenated features to predict HN. For patients who require routine check-ups, kidney ultrasounds are taken on each visit, resulting in a relatively short sequence of ultrasound images over time. The current approach, however, cannot make a prediction on a sequence of ultrasound images; it is designed to make a prediction based solely on a pair of ultrasound images taken at one time point. As a result, data collected from previous hospital visits are left unused, foregoing potentially valuable temporal information from changes between visits.

We seek to understand if incorporating ultrasounds obtained over multiple hospital visits improves the prediction of obstructive HN. Video-based modeling draws many parallels to the current problem. In action recognition, the task is to classify an action given a sequence of image frames. Similarly, the task is to classify renal obstruction given a sequence of ultrasound images. Since each patient may differ in their number of hospital visits, we would like that the model be flexible to handle variable-length input to reflect this. We focus particularly on methods from action recognition that build from 2D CNNs for two reasons: (1) examples in training data typically contain few hospital visits (most frequently three visits), and (2) it allows pretraining from the 2D Siamese CNN from the original study. These methods typically require some form of temporal fusion of convolutional features \cite{review_action}. 

Thus, we adapt the original single-visit model to perform multi-visit inference through temporal fusion methods (convolutional pooling, long short term memory, temporal shift module) to determine if incorporating data from past hospital visits improves prediction of renal obstruction. We show that all multi-visit methods fail to perform better than the original single-visit model. Additionally, we show that prediction using data from the latest visit is no different than predicting using data from the first visit. In a clinical setting, these findings provide a positive outlook on the flexibility of the single-visit model with regards to the availability of patient ultrasound data.

\section{Methods}
\subsection{Study Design}

Research ethics board approval was acquired for image collection from each study site and data sharing agreements were obtained for secure transfer to SickKids. Two fixed-plane (transverse and sagittal) images were taken for the left and right kidney of each patient. Each kidney is considered independent from one another. Ultrasounds taken over multiple visits make up a sequence of two ultrasound views for a left/right kidney. Images were preprocessed as in the original study \cite{hydro}. Center cropping was done to remove text annotations and ultrasound beam borders. Contrast limited adaptive histogram equalization is used to normalize the contrast of the images. Lastly, images were resized to 256x256 pixels. Code is made available at: \href{https://github.com/stan-hua/temporal_hydronephrosis}{GitHub}.

For training, ultrasound imaging data were sourced from 
the Hospital for Sick Children (SickKids). In addition to the internal test set, we evaluated models on recent data collected from the same institution and data from two external institutions: 
Lucile Packard Children's Hospital (Stanford)
Children's Hospital of Philadelphia (CHOP).

\subsection{Evaluation Metrics}

Area under the receiver operating characteristic (AUROC) and area under the precision-recall curve (AUPRC) were used to compare the performance of various models on the binary classification task. AUROC provides a summary of how well the model can distinguish between the classes, but it can be easily inflated if there is class imbalance. On the other hand, AUPRC describes how well the model identifies positive patients without false positives \cite{auroc_auprc}. Estimated using bias-corrected and accelerated (BCa) bootstrap, 95\% confidence intervals are reported for each metric \cite{bootstrap}. AUPRC and AUROC were calculated via the \texttt{torchmetrics} package, and bootstrap was performed using the \texttt{arch} package.

\subsection{Baseline model}

The model from the original paper is treated as the baseline. It is a Siamese convolutional network architecture that takes in 2 kidney ultrasound images (transverse and sagittal plane) and predicts renal obstruction. It consists of 7 convolutional layers and 3 fully-connected layers (See Figure 2). The two ultrasound images are each fed into seven convolutional layers and one linear layer. The outputs are then concatenated and passed through two more linear layers and a softmax for binary classification. 

The baseline model was retrained using the described training split. As in the original paper, ultrasounds from a patient's hospital visits were considered independent during training. We compared test set results from baseline prediction on the latest visit versus the first visit. When comparing the baseline against other methods, inference on ultrasounds from the latest hospital visit is reported. 

A randomized grid search with 5-fold cross-validation on the training set was performed to find optimal hyperparameters for the baseline model. Hyperparameters  with the best average validation AUPRC across folds were kept. Negative log-likelihood loss was optimized via Stochastic Gradient Descent (SGD) with a learning rate of 0.005, momentum of 0.9, weight decay of 0.0005, and a batch size of 16.

\begin{figure}
\centering
\includegraphics[width=3.5in]{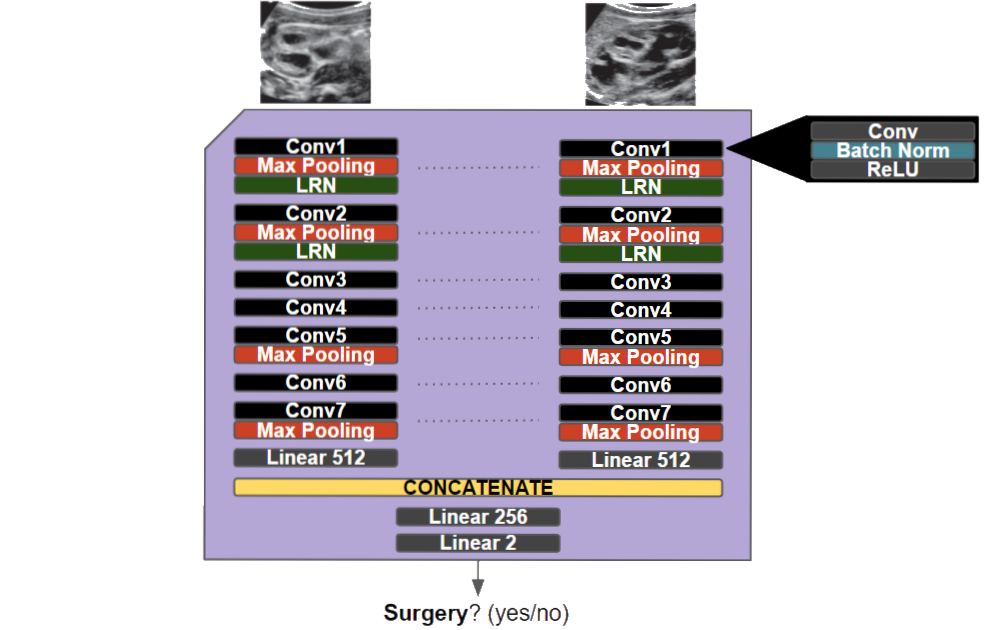}
\caption{Diagram of siamese baseline model.} \label{fig2}
\end{figure}

\subsection{Methods to Adapt Single-visit Models} 

We wanted to determine if ultrasounds taken over multiple hospital visits may contain spatio-temporal information that could improve discrimination of obstructive HN. We explored the following methods to adapt the baseline model to handle multi-visit inference: (1) Average Prediction, (2) Convolutional Pooling, (3) Temporal Shift Module and (4) Long Short-Term Memory. The first three methods do not require retraining, as they only modify the baseline model's forward pass. As a result, they come at no cost to the model's parameter size.

As the number of hospital visits varies with each example, this corresponds to a varying image-sequence length. During training, validation and testing, each example was fed into the model one at a time. Gradients were accumulated over examples to test for different effective batch sizes. Similar to the baseline, a randomized grid search for hyperparameters was done when training models with each of the following methods. Models were implemented in \texttt{pytorch-lightning} and trained on an NVIDIA RTX 3080 GPU.


\subsubsection{Average Prediction} (Avg. Pred) is a naive extension to the single-visit baseline model, in which prediction is performed for each available time point and the predictions are averaged over time. The single-visit model is used to predict on ultrasounds from each hospital visit and mean is taken over the logit outputs. The mean logits are then passed into a softmax as in the single-visit model.

\subsubsection{Convolutional Pooling} (Conv. Pooling) has been effective in video classification tasks, performing significantly better than single-frame models, and early 3D convolutional neural networks (CNN) \cite{conv_pooling,c3d}. It is a pooling method to aggregate convolutional image features across time, where a max operation is performed on features extracted from the last convolutional layer. In the adapted forward pass, the 1024-dimensional feature vectors (from concatenating the outputs of the Siamese layers) are extracted for each time point. A max operation is done on extracted features over time, and the resulting 1024-dimensional feature vector is passed through the remaining layers.

\subsubsection{Temporal Shift Module} (TSM) is a more recently proposed method for temporal fusion that showed great performance on action recognition benchmarks \cite{tsm}. Built on top of 2D CNNs, TSM allows modeling of temporal dynamics through ``temporal shifts" on feature maps between convolutional layers. 

Between convolutional layers, the output of the previous layer is a feature map of the form (T, C, H, W), where T is the number of time points, C is the number of channels, H is the height and W is the width. A portion of the feature map's channels (C) is shifted once forward across time, and a non-overlapping portion is shifted once backwards across time. Adapting the single-visit model, temporal shifts occur between the Siamese convolutional layers, which means each ultrasound view (sagittal and transverse) is shifted independently across time. After the convolutional layers, features from the latest time point are fed into the remaining layers of the model to produce a single prediction.

\subsubsection{Long Short Term Memory} (LSTM) is a variant of recurrent neural network containing three gates and a memory cell that allow for the updating, forgetting, and persistence of hidden states across time steps. Hybrid architectures combining both convolutional networks and recurrent networks have been particularly useful in the modeling of spatio-temporal data (e.g., video, ECG) \cite{liao,zihlmann,stanley,shi,sharma}. As with Conv. Pooling, for the LSTM 1024-dimensional feature vectors are extracted at each time point. The sequence of feature vectors are then fed to a bidirectional LSTM, and the final hidden representation is passed to the remaining layers of the model.

\section{Results}

\subsection{Data}

SickKids data was randomly shuffled and 70\% (321 patient trajectories of differing length, from 600 clinic visits total) were placed in the training set, while the remaining (80 patient trajectories, 153 clinic visits) were left for evaluation. In contrast to the previous study, we did not split patients by time, so as to increase the number of patients in the test set with 2 or more hospital visits. In addition to this random split, we included 3 other test sets to better understand the performance of the models tested.

 An additional test set from SickKids (202 patients, 244 examples), termed ``silent trial" data, was collected more recently to evaluate the baseline model. To avoid data leakage, none of the silent trial patients were in the internal data used for the training and testing. From the silent trial, 12\% of the examples were positive. Data collected at Stanford contained 102 patients (204 examples) most with two or more visits, but there are relatively few positive examples (5\%). Meanwhile, data collected at CHOP included 89 patients (89 examples), where each patient was imaged only once, and positive examples constitute a greater proportion of the data (67\%).

\begin{figure}
\centering
\includegraphics[width=3in]{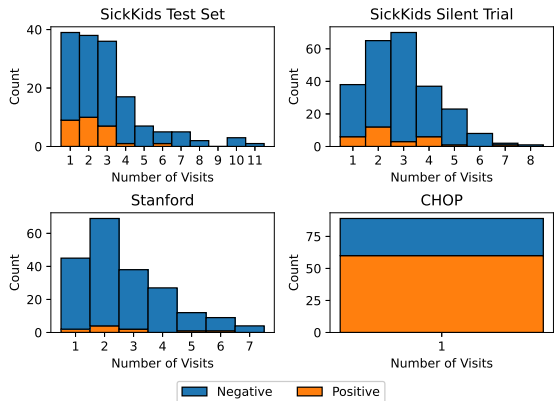}
\caption{Stacked bar plots describe the number of hospital visits for examples in the test sets. Blue and orange represent the negative and positive examples, respectively.} \label{fig1}
\end{figure}

\subsection{Performance of Baseline Model on a Patient's First versus Latest Ultrasounds}

The first experiment tested if the single-visit inference performs significantly better with data from a patient's latest hospital visit, compared to data from their first visit. The AUROC and AUPRC are reported for each approach. We evaluated the two approaches on the test sets. Since CHOP only contains single-visit examples, it is excluded. Evaluation on the test sets demonstrate no significant difference in performing inference using the first visit versus the last visit (Table \ref{tab:tab1}).

\begin{table}[t]
\centering
\caption{Single-visit baseline model's performance given ultrasounds from a patient's first hospital visit versus their latest hospital visit. 95\% bootstrapped confidence intervals are provided in brackets.}
\label{tab:tab1}
\resizebox{12cm}{!}{%
    \begin{tabular}{lc|llll}
    \multicolumn{1}{c}{Dataset} & \multicolumn{1}{l|}{Model} & \multicolumn{1}{c}{AUROC}                    &  &  & \multicolumn{1}{c}{AUPRC}                    \\ \hline
    \multirow{2}{*}{SickKids Test Set}     & First & 89.54 {[}79.88, 94.61{]} &  &  & 65.86 {[}43.63, 80.63{]} \\
            & Latest                     & 92.69 {[}84.41, 96.73{]} &  &  & 74.06 {[}49.78, 87.41{]} \\ \hline
    \multirow{2}{*}{SicKKids Silent Trial} & First & 93.04 {[}87.29, 96.10{]} &  &  & 56.49 {[}36.12, 72.10{]} \\
            & Latest                     & 95.59 {[}91.62, 97.86{]} &  &  & 71.10 {[}45.71, 84.93{]} \\ \hline
    \multirow{2}{*}{Stanford}              & First & 86.65 {[}54.89, 99.43{]} &  &  & 69.94 {[}31.45, 92.20{]} \\
            & Latest                     & 86.29 {[}57.37, 97.90{]} &  &  & 67.61 {[}29.49, 90.87{]}
    \end{tabular}%
}
\end{table}

\subsection{Performance of Single-Visit vs. Multi-Visit Models}
Primarily, we wanted to examine if incorporating ultrasounds from all hospital visits aids in the discrimination of renal obstruction. We compared the performance of the baseline model and each of the methods (avg. pred., conv. pooling, LSTM, TSM) on the test sets. Three of the 4 described methods only require modification of the forward pass of the single-visit baseline model without additional parameters. This allowed us to directly use the weights from the retrained baseline model. We include the results from adapting the pretrained baseline, as well as results from retraining the model with multi-visit examples. As all multi-visit models fail to produce AUROC/AUPRC values outside of the 95\% confidence interval of the baseline model, they fail to significantly outperform the single-visit baseline model (Table \ref{tab:tab2}). In some cases, the trained model with avg. pred. or TSM perform significantly worse than the baseline. Additionally, training multi-visit models with conv. pooling or LSTM does not hurt single-visit inference as seen in the performance on CHOP, which contains examples with only 1 visit each.

\begin{table}[t]
\centering
\caption{Comparison of model performance of single-visit versus multi-visit models. `(Pretrained)' denote using weights of pretrained baseline model with modification to perform multi-visit inference. As CHOP data is cross-sectional, adapted pretrained single-visit models are not included since these models default to the single-visit model when predicting on single-visit patients. 95\% bootstrapped confidence intervals are provided in brackets. Top models according to AUROC and AUPRC for each test set are indicated with bold typeface}
\label{tab:tab2}
\resizebox{\textwidth}{!}{%
\begin{tabular}{ll|llll}
\multicolumn{1}{c}{Dataset}            & \multicolumn{1}{c|}{Model}          & \multicolumn{1}{c}{AUROC} &  &  & \multicolumn{1}{c}{AUPRC} \\ \hline
\multirow{8}{*}{SickKids Test Set}     & Baseline                            & 92.69 {[}84.41, 96.73{]}  &  &  & 74.06 {[}49.78, 87.41{]}  \\
                      & (Pretrained) Avg. Prediction & 92.20 {[}82.65, 96.67{]} &  &  & 74.54 {[}50.29, 87.48{]} \\
                      & (Pretrained) Conv. Pooling   & 94.03 {[}87.89, 97.40{]} &  &  & 75.60 {[}51.69, 88.41{]} \\
                      & (Pretrained) TSM             & 92.26 {[}79.95, 96.60{]} &  &  & 75.55 {[}54.66, 87.71{]} \\
                      & Avg. Prediction              & 85.23 {[}74.96, 91.67{]} &  &  & 59.17 {[}37.84, 75.25{]} \\
                      & \textbf{Conv. Pooling}       & \textbf{94.86} {[}89.60, 97.69{]} &  &  & \textbf{78.60} {[}57.63, 90.38{]} \\
                      & LSTM                         & 93.89 {[}87.49, 97.19{]} &  &  & 78.56 {[}58.86, 89.79{]} \\
                      & TSM                          & 84.06 {[}71.22, 91.79{]} &  &  & 68.17 {[}47.38, 81.65{]} \\ \hline
\multirow{8}{*}{SickKids Silent Trial} & Baseline                            & 95.59 {[}91.62, 97.86{]}  &  &  & 71.10 {[}45.71, 84.93{]}  \\
                      & (Pretrained) Avg. Prediction & 97.32 {[}94.64, 98.78{]} &  &  & 72.81 {[}49.02, 87.39{]} \\
                                      & \textbf{(Pretrained) Conv. Pooling} & \textbf{97.63} {[}95.17, 98.96{]}  &  &  & 73.73 {[}50.20, 88.19{]}  \\
                      & (Pretrained) TSM             & 92.04 {[}82.51, 96.65{]} &  &  & 70.98 {[}48.31, 84.38{]} \\
                      & Avg. Prediction              & 93.15 {[}86.10, 96.72{]} &  &  & 71.44 {[}50.90, 84.65{]} \\
                      & \textbf{Conv. Pooling}       & 96.76 {[}94.03, 98.45{]} &  &  & \textbf{76.73} {[}56.33, 88.65{]} \\
                      & LSTM                         & 96.57 {[}93.74, 98.28{]} &  &  & 68.48 {[}46.98, 83.07{]} \\
                      & TSM                          & 96.20 {[}92.49, 98.12{]} &  &  & 74.13 {[}53.56, 87.02{]} \\ \hline
\multirow{8}{*}{Stanford}              & Baseline                            & 86.29 {[}57.37, 97.90{]}  &  &  & 67.61 {[}29.49, 90.87{]}  \\
                      & (Pretrained) Avg. Prediction & 86.39 {[}49.57, 99.43{]} &  &  & 72.99 {[}36.17, 94.14{]} \\
                      & (Pretrained) Conv. Pooling   & 88.45 {[}55.13, 99.25{]} &  &  & 70.40 {[}32.24, 92.06{]} \\
                      & \textbf{(Pretrained) TSM}    & 86.60 {[}54.80, 99.59{]} &  &  & \textbf{73.83} {[}36.20, 94.77{]} \\
                      & Avg. Prediction              & 80.77 {[}53.83, 97.80{]} &  &  & 58.13 {[}20.04, 84.79{]} \\
                      & Conv. Pooling                & 85.00 {[}48.28, 99.43{]} &  &  & 71.04 {[}34.17, 93.21{]} \\
                      & \textbf{LSTM}                & \textbf{91.49} {[}70.55, 99.48{]} &  &  & 70.73 {[}32.45, 92.67{]} \\
                      & TSM                          & 85.00 {[}51.17, 99.17{]} &  &  & 61.13 {[}22.65, 87.67{]} \\ \hline
\multirow{5}{*}{CHOP} & Baseline                     & 90.34 {[}82.56, 95.18{]} &  &  & 95.41 {[}90.43, 97.87{]} \\
                      & Avg. Prediction              & 78.10 {[}65.05, 87.42{]} &  &  & 84.42 {[}68.87, 92.39{]} \\
                      & \textbf{Conv. Pooling}       & \textbf{91.26} {[}82.06, 96.05{]} &  &  & 95.39 {[}88.29, 98.16{]} \\
                      & \textbf{LSTM}                & 90.75 {[}82.92, 95.50{]} &  &  & \textbf{96.05} {[}91.87, 98.21{]} \\
                      & TSM                          & 82.18 {[}71.34, 89.62{]} &  &  & 90.99 {[}80.48, 95.47{]}
\end{tabular}%
}
\end{table}



\section{Discussion}

Previous models of HN use ultrasound images from only a single visit \cite{hydro}. In practice, HN patients may have been imaged several times over repeated hospital visits, in order to observe for obstruction. This raised two questions: (1) Is it important to predict on data from the latest time point?, and (2) Can previously collected ultrasounds improve model prediction of renal obstruction?  This work shows that longitudinal modeling of HN ultrasounds offers minimal benefits to predictive performance, largely because prediction from a single clinical visit is so strong. Specifically, the SickKids test and Silent Trial data show convolutional pooling is the best performing method but no method is significantly different, aside from TSM and average prediction which are significantly lower performing. 

\textbf{Limitations:} While the results of this work in the SickKids data show independent, replicated results, sample sizes are too small in the Stanford data to distinguish performance between methods. Moreover, CHOP data only had observations from a single clinical visit. Therefore, data in institutions beyond SickKids are currently too limited to understand whether a single visit's ultrasound data is sufficient for clinically relevant prediction at hospitals beyond SickKids. 

\textbf{Future work:} Future work will consider the convolutional pooling model as the best way to integrate ultrasound data across HN patient visits. However, where multi-visit data is not available, this work suggests that this is not a major limitation and modeling based on a single visit can be extremely powerful. 

\textbf{Conclusions:} Experiments on two internal and two external test sets show that (1) there is no significant difference in single-visit prediction using ultrasounds from a patient's first or latest hospital visit, and (2) incorporating ultrasounds from a patient's previous hospital visits does not significantly improve prediction of renal obstruction over using the latest (most recent) ultrasound. In a clinical setting, these results imply that the single-visit model would be sufficient in aiding diagnosis of obstructive hydronephrosis, given kidney ultrasounds from any one of the patient's hospital visits. 

\newpage

\bibliography{report} 
\bibliographystyle{spiebib} 

\end{document}